\documentclass[conference]{IEEEtran}
\usepackage{times}

\usepackage{cite}
\usepackage{amsmath,amssymb,amsfonts}
\usepackage{graphicx}
\usepackage{textcomp}
\usepackage{xcolor}
\usepackage[numbers]{natbib}
\usepackage{multicol}
\usepackage{multirow}
\usepackage{bm}
\usepackage{algorithm}
\usepackage{algpseudocode}
\usepackage{todonotes}
\usepackage{afterpage}
\usepackage[caption=false]{subfig}

\usepackage[bookmarks=true]{hyperref}
\def\BibTeX{{\rm B\kern-.05em{\sc i\kern-.025em b}\kern-.08em
    T\kern-.1667em\lower.7ex\hbox{E}\kern-.125emX}}

\renewcommand{\vec}[1]{\bm{#1}}
\newcommand{\mat}[1]{\mathrm{\bm{#1}}}

\begin{document}

\title{MultiSCOPE: Disambiguating In-Hand Object Poses with Proprioception and Tactile Feedback}

\author{Author Names Omitted for Anonymous Review. Paper-ID 386}
\author{
\IEEEauthorblockN{Andrea Sipos}
\IEEEauthorblockA{\textit{Robotics Department} \\
\textit{University of Michigan}\\
Ann Arbor, USA \\
asipos \textit{at} umich \textit{dot} edu}

\and

\IEEEauthorblockN{Nima Fazeli}
\IEEEauthorblockA{\textit{Robotics Department} \\
\textit{University of Michigan}\\
Ann Arbor, USA \\
nfz \textit{at} umich \textit{dot} edu}
}

\maketitle

\begin{abstract}
In this paper, we propose a method for estimating in-hand object poses using proprioception and tactile feedback from a bimanual robotic system. Our method addresses the problem of reducing pose uncertainty through a sequence of frictional contact interactions between the grasped objects. As part of our method, we propose 1) a tool segmentation routine that facilitates contact location and object pose estimation, 2) a loss that allows reasoning over solution consistency between interactions, and 3) a loss to promote converging to object poses and contact locations that explain the external force-torque experienced by each arm. We demonstrate the efficacy of our method in a task-based demonstration both in simulation and on a real-world bimanual platform and show significant improvement in object pose estimation over single interactions. Visit \url{www.mmintlab.com/multiscope/} for code and videos.
\end{abstract}

\section{Introduction}

Dexterous object manipulation is a challenging open problem in robotics. Inaccurate in-hand pose estimation of grasped objects is an important cause of failure during robust robot manipulation. Inaccurate pose estimates can lead to misalignment (that can cause slip), unforeseen contacts, and jamming. These failure modes result in unreliable interactions between the robot and its environment when performing tasks such as tool use and assembly.

Many current state-of-the-art approaches to this problem use visual feedback to identify objects in the scene and estimate their poses \cite{kokic2019learning, tremblay2018deep, tremblay2020indirect, wen2020robust, romero2013non, du2021vision, honda1998real, haidacher2003estimating, li2020hand, paul2021object}. These approaches produce poor or unreliable estimates 
in vision-deprived environments. Vision deprivation occurs naturally due to occlusions from the robot or environment during task execution (e.g., during fine assembly or operation in narrow spaces) and is often unavoidable \cite{zhong2021tampc, zhong2022soft, van2022learning, wi2022virdo}. Recent progress in high-resolution collocated tactile sensing has enabled in-hand pose estimation \cite{villalonga2021tactile, ma2021extrinsic, yamaguchi2019tactile, bimbo2016hand, zwiener2018contact, alvarez2017tactile, oller2022membranes}. While these novel sensors present many new possibilities for tactile feedback, they also present several challenges including high cost, components that easily fatigue, difficult to model dynamics between the robot and grasped object, and data processing complexity. 

\begin{figure}
    \centering
    \includegraphics[width=0.45\textwidth]{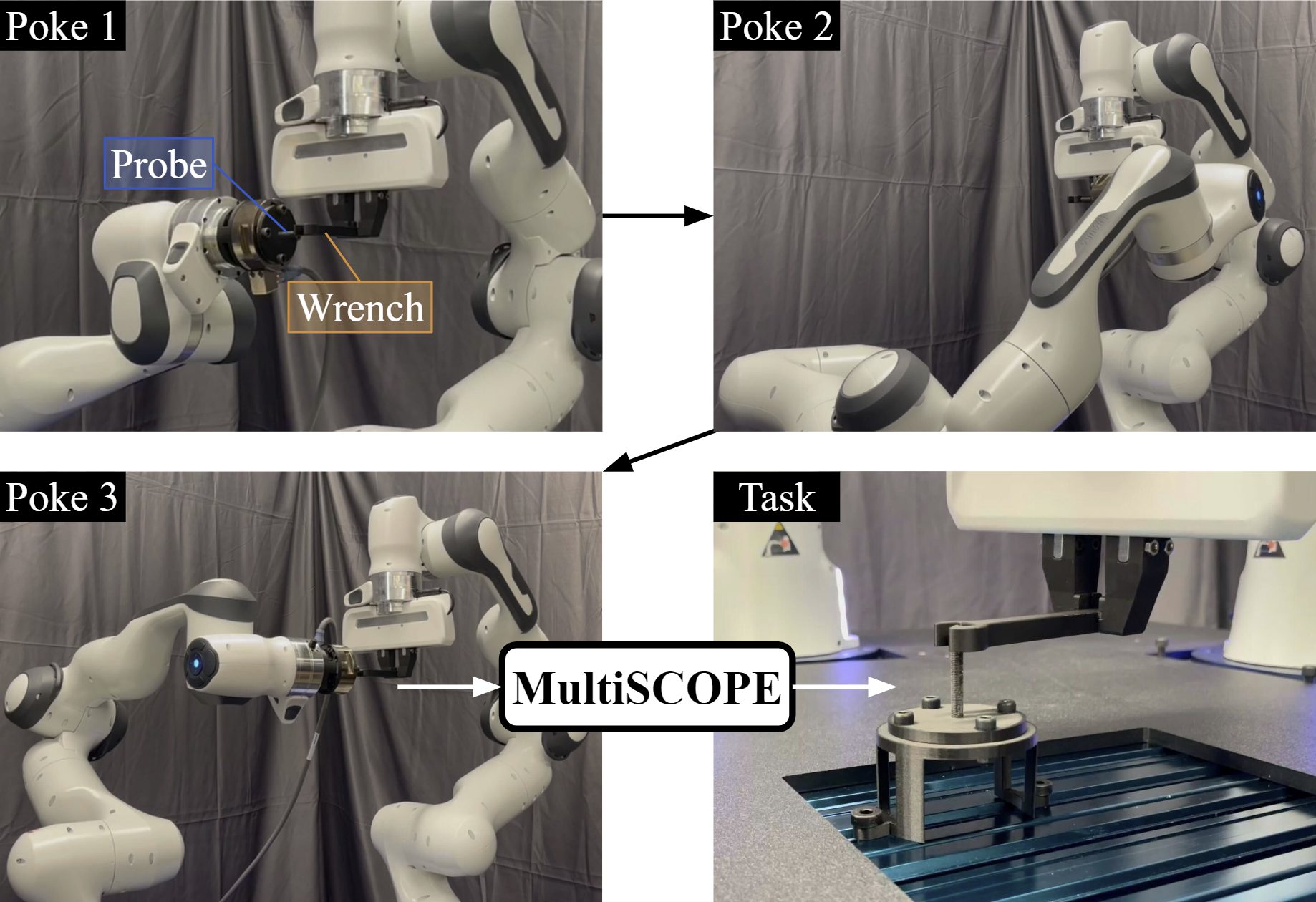}
    \caption{Each robot in this bimanual system is grasping an object (probe and wrench) in unknown configurations. Using our method, MultiSCOPE, the robot is able to localize both simultaneously using a sequence of information-gathering "poking actions", its own proprioception and external wrench measurements, and object models. After running MultiSCOPE to estimate object poses, the robot will attempt to use the wrench tool.}
    \label{fig:teaser}
\end{figure}

Here, we present MultiSCOPE: a method to simultaneously estimate the poses of two objects grasped in unknown configurations by two collaborative arms using only proprioception and 6-DOF force-torque sensing at the wrists, shown in Fig.~\ref{fig:teaser}. Our method works by iteratively bringing the two objects into contact with one another and uses two complementary particle filters to accurately estimate in-hand object poses. These filters exploit mutual information between known object geometries and detected contact events. Specifically, we contribute:
\begin{itemize}
    \item A framework for sequential bimanual interactions to reduce pose ambiguity through iterative Bayesian filtering,
    \item A memory loss to promote consistency between sequential information-gathering physical interactions,
    \item A tool segmentation algorithm to improve object pose and contact location estimation, and
    \item A wrench loss to promote converging to object poses and contact locations that explain the external force-torque experienced by each arm.
\end{itemize}
We demonstrate the efficacy of our proposed approach on simulated and real data using 2 Franka Emika Panda robots. We emphasize that our approach does not require visual feedback or high-resolution tactile sensing and is compatible with most existing robot hardware.

\section{Related Work}
\label{sec:related_work}

In-hand object pose estimation is an important challenge in robotics. The extensive literature on the subject can be grouped into two approaches: vision-based and tactile. More recently, there is an increasing amount of overlap in visuotactile approaches that we discuss below: 
\subsection{Vision-Based State Estimators}
Many works on in-hand object pose estimation use vision-based approaches \cite{honda1998real, haidacher2003estimating, li2020hand}. Vision-based methods are significantly impacted by occlusion. Aside from occlusion, two main challenges for these methods are i) multi-object interactions (i.e. cluttered scenes) \cite{wen2020robust, romero2013non} and ii) reasoning about contact \cite{tremblay2018deep, tremblay2020indirect, du2021vision}. One of the main advantages of these methods is that they often mitigate the need for object models via learning because vision is a much more global sense than touch \cite{tremblay2018deep, tremblay2020indirect, paul2021object}. 

\subsection{Tactile and Visuotactile-Based State Estimators}
These works use a variety of tactile sensors and visuotactile methods to reason about pose at the interface between the robot and the object. There are many different types of tactile and visuotactile sensors including robotic skin \cite{fan2022enabling}, tactile fingertips \cite{lepora2022digitac, bimbo2016hand, yamaguchi2019tactile}, Soft Bubbles \cite{zhong2022soft}, and GelSlim \cite{villalonga2021tactile, ma2021extrinsic} that are being used in robotic manipulation research. While these sensors hold a lot of promise, they are often expensive and fragile. In our method, we use robot-estimated wrenches and an off-the-shelf force-torque sensor to gain tactile feedback. These methods are more accessible than most current tactile and visuotactile sensors. Two other works that also use robot proprioception to estimate contact are \cite{manuelli2016localizing, zwiener2018contact}. However, these methods estimate contact on the surface of the robot rather than an object grasped by the robot. \cite{sipos2022scope} extends this idea to an object grasped by a robot, as well as to objects grasped in a bimanual system. Several other methods use finger/robot position combined with RGB(-D) data to estimate object shape and pose \cite{kokic2019learning, alvarez2017tactile, saund2021clasp}. Additionally, several works have explored using particle filtering methods \cite{manifoldpf, probingpf, rbpf, hapticrender, contactformations} to estimate contact between a robot and its environment. Generally, these methods consider single robot arms or known in-hand object pose but an uncertain environment. Our method could reasonably be integrated with the above vision-based and tactile techniques discussed here.

\section{Methodology}
\label{sec:methodology}
In this paper, we extend recent work from \citet{sipos2022scope} on single-action object pose estimation with proprioception and tactile feedback called Simultaneous Contact and Object Pose Estimation, or SCOPE. Our method extends this framework to multiple contact-rich interactions in order to achieve object pose estimates that are consistent across actions and result in more accurate object pose estimates. We emphasize that in this framework, the robot is estimating the pose of both grasped objects (e.g., the probe and tool in Fig.~\ref{fig:teaser}).

\subsection{Assumptions}
We assume that the grasped objects are rigid and that we have access to their geometric models. We further assume that the robot is capable of estimating externally applied wrenches. This is a common functionality for most existing cobots (e.g., Franka Emika Panda and Kuka lbr iiwa) and can be implemented using an external force-torque sensor at the wrist for arms without this capability. We also assume that we have access to robot proprioception. We assume that contacts are well-approximated by the point contact model and that no moments are transmitted through the contact location. Finally, we assume that the robot's exploratory actions (pokes) result in contact with no relative slip. These assumptions are the same as SCOPE \cite{sipos2022scope} and no additional information is required.

\subsection{SCOPE Review}
SCOPE \cite{sipos2022scope} leverages two complementary particle filters to jointly estimate the contact location and grasped object poses (for two objects grasped in unknown configurations) using robot proprioception and tactile feedback from a single interaction. The first filter is called the Contact Particle Filter for Grasped Objects, or CPFGrasp. For each arm, this filter takes as input the measured robot wrench at the end-effector $\vec{\Gamma}_{EE} \in \mathbb{R}^6$ and an estimate of the pose of the grasped object represented as a transform between the object and end-effector frames $\mat{H}_O \in \mathbb{SE}(3)$. In this work, we estimate grasped object poses in $\mathbb{SE}(2)$ because the pose is constrained by the parallel jaw grippers of the Frankas. The output of this filter is a belief distribution over the estimated contact locations on the objects' surfaces and the forces applied at these locations conditioned on $(\vec{\Gamma}_{EE}, \mat{H}_O)$. This distribution is approximated by a set of particles referred to as the Contact Location Particles (CLPs): $\mat{R}_c = \{\vec{r}_i \in \mathbb{R}^3 \; | \; i \in \mathbb{N} < n_{clp} \}$ where $\vec{r}$ denotes the contact location w.r.t. to the end-effector frame and where $n_{clp}$ denotes the number of particles. Once converged, CPFGrasp is able to provide a score $\mat{S} = \{ s_i \in \mathbb{R} \; | \; i \in \mathbb{N} < n_{clp} \}$ for how well an estimated pose is able to explain the measured robot wrenches:
\begin{align*}
    (\mat{R}_c, \mat{S}) = \text{CPFGrasp}(\mat{H}_O, \Gamma_{EE})
\end{align*}

The second filter, SCOPE, maintains a belief distribution over object poses and updates this belief distribution using the scores computed by CPFGrasp. This distribution is approximated using the Object Pose Particles (OPPs): $\mat{H} =\{ \mat{H}^i_O \in \mathbb{SE}(2) \; | \; i \in \mathbb{N} < n_{pp}\} $ where $n_{pp}$ denotes the number of particles. Given an initial distribution of object pose particles, CPFGrasp provides scores for how well each particle explains the measured wrench, and SCOPE updates these particles based on their computed scores. These two filters are run sequentially until convergence. Each object in the bimanual system has its own OPP distribution. OPPs from each grasped object in the bimanual interaction are scored in pairs using the losses described below.

\vspace{3pt}
\noindent\textbf{Losses} The losses introduced in SCOPE are the penetration loss, force alignment loss, and contact consistency loss. Penetration loss penalizes OPP pairs that are in collision. Contact consistency loss penalizes OPP pairs that have spatially distant contact locations. Finally, force alignment loss penalizes OPP pairs that have misaligned applied force. These losses are formalized in Eqs.~\ref{eq:pen_loss}, ~\ref{eq:contact_loss}, and ~\ref{eq:align_loss}:
\begin{align}
    \label{eq:pen_loss}
    \mathcal{L}_P &= \max\{0, N_{pp} - \epsilon_{pp}\} \\
    \label{eq:contact_loss}
    \mathcal{L}_C &= \sum s_{t}s_{p} \| \vec{r}_{t} - \vec{r}_{p} \|_2 \\
    \label{eq:align_loss}
    \mathcal{L}_F &= \sum s_{t}s_{p} \| -\vec{w}_t - \vec{w}_p \|_2
\end{align}
where we introduce the subscripts $t$ and $p$ for clarity of notation to describe the objects (e.g., tool and probe). Here, $N_{PP}$ is the number of points in penetration for a given OPP pair and $\epsilon_{PP}$ is a threshold for penetration in the ground truth. $\vec{r}_t$ and $\vec{r}_p$ are the CLPs for the tool and probe respectively. $n_{clp}$ is the number of CLPs for each of the objects. $s_{t}$ and $s_{p}$ are the scores of each CLP, and $\vec{w}_t$ and $\vec{w}_p$ are the external force estimates at each contact location in the world frame. The summations are taken pairwise.

\subsection{MultiSCOPE}
\label{sec:multiscope}
To extend SCOPE for sequential contacts, we introduce a tool segmentation algorithm that improves contact location initialization and convergence properties, a memory loss to allow our method to find temporally consistent solutions, and a wrench loss to promote consistency in wrench measurements. In the following, we provide the details of each contribution and summarize our proposed method in Algs.~\ref{alg:SCOPE} and ~\ref{alg:MultiSCOPE}. Alg.~\ref{alg:SCOPE} contains extensions to the algorithm presented in \cite{sipos2022scope}.

\begin{algorithm}
\caption{Tool Segmentation}
\label{alg:tool_seg}
\begin{algorithmic}
\Procedure{SegmentTool}{$\mathbf{n}_{\mathcal{S}}, n_{clusters}, \varepsilon, n_{min}$}
\State $\mathbf{n}_{unique}$ $\gets$ getUnique($\mathbf{n}_{\mathcal{S}}$)
\State centroids $\gets$ KMeans($\mathbf{n}_{unique}, n_{clusters}$)
\State groups$_{\mathbf{n}}$ $\gets$ groupNorms($\mathbf{n}_{\mathcal{S}}$, centroids)
\State faces $\gets$ DBSCAN(groups$_{\mathbf{n}}, \varepsilon, n_{min}$)
\State \textbf{return} faces
\EndProcedure
\end{algorithmic}
\end{algorithm}

\subsubsection{Tool Segmentation}
\label{sec:tool_segmentation}

\begin{figure}
    \centering
    \includegraphics[width=0.48\textwidth]{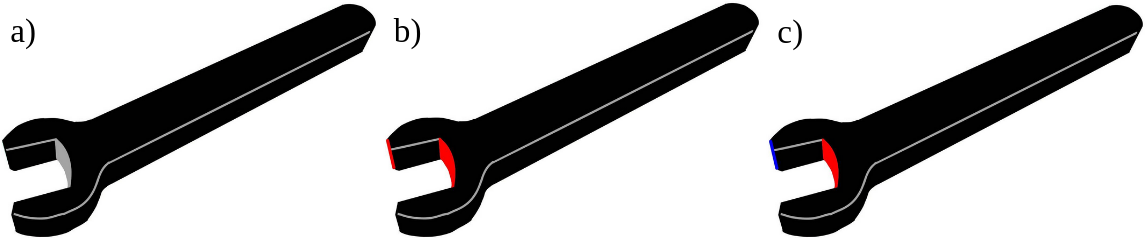}
    \caption{a) Filled gray: useful face of the wrench tool, also one of the smaller faces b) Filled red: faces of the wrench that share the same surface normal c) Filled red and blue: faces that share the same surface normal segmented into different faces based on spatial proximity.}
    \label{fig:wrench_seg}
\end{figure}

\begin{figure*}
    \centering
    \includegraphics[width=\textwidth]{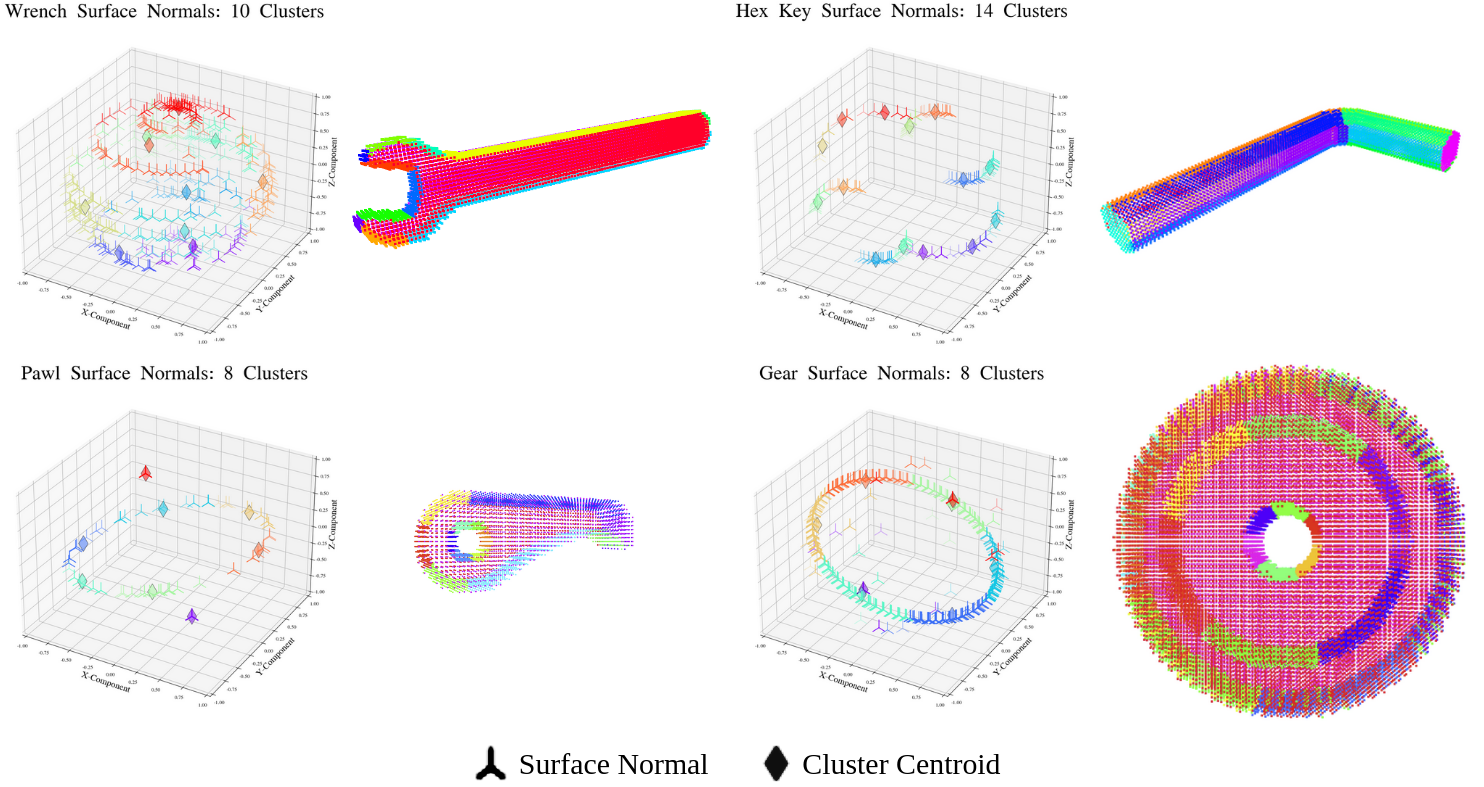}
    \caption{The result of K-Means clustering of unique surface normals alongside the resulting tool segmentation for each of the wrench, hex key, pawl, and gear tools. Each tri-mark is the 3D coordinate representing the X-, Y-, and Z-component of the surface normal unit vector. In the resulting tool segmentation, each face is distinctly colored.}
    \label{fig:tool_segmentation}
\end{figure*}

In order to ensure better convergence properties from the CPFGrasp introduced in \cite{sipos2022scope}, we propose a tool segmentation method outlined in Alg.~\ref{alg:tool_seg} that is designed to improve CLP initialization and consequently pose estimation. The original CPFGrasp method uniformly sampled from the tool surface, and this means that smaller faces are less likely to be sampled than large ones. However, in the case of the grey face in Fig.~\ref{fig:wrench_seg}a, these faces may be intended to make contact with the environment. If there are no CLPs on this face at initialization, it is unlikely that the filter will converge to it because the particles would have to traverse high-error faces to get there. To address this, we propose a tool segmentation algorithm that samples $N_{face}$ particles on each face, ensuring better convergence properties for CPFGrasp. In order to do this, we segment the tool into faces, then sample points from each. We define a face as a group of points that have similar surface normals and are spatially close together. To find groups of similar surface normals, we use K-Means clustering \cite{scikit-learn} of unique surface normals. After grouping by surface normal, we use DBSCAN \cite{o3dDBSCAN} to further separate faces that share the same surface normal but are spatially distant from one another, an example of which is shown in Fig.~\ref{fig:wrench_seg}b-c. Tool segmentations produced by this method for all tools are shown in Fig.~\ref{fig:tool_segmentation}. This method's parameters are easy to modify for new tools. This contribution improves the performance and reliability of MultiSCOPE over random CLP initialization, as we show in Fig.~\ref{fig:toolseg_comp}.

\begin{figure}[b]
    \centering
    \includegraphics[width=0.48\textwidth]{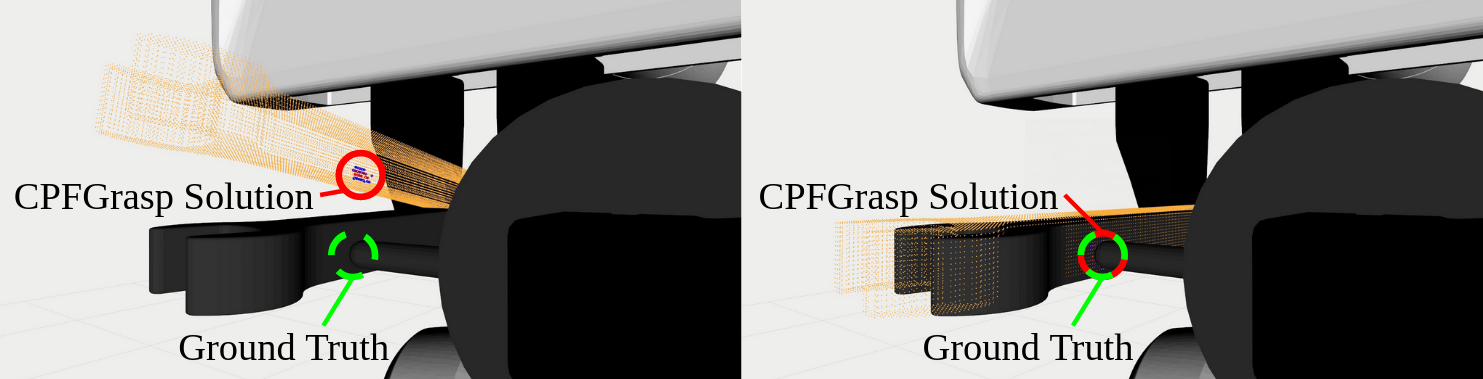}
    \caption{Left: The CPFGrasp solution does not resolve the wrist wrench experienced by the robot. Right: The CPFGrasp solution is able to resolve the wrist wrench. $\mathcal{L}_{\Gamma}$ allows us to distinguish between these cases.}
    \label{fig:ft_loss_example}
\end{figure}

\subsubsection{Wrench Loss}
While the penetration, force alignment, and contact consistency losses in \cite{sipos2022scope} encode many properties of contact-rich interactions, they don't explicitly enforce that the algorithm converges to the object pose(s) that best explains the wrench the system is experiencing at the end-effector $\Gamma_{EE}$. To this end, we add a wrench loss $\mathcal{L}_{\Gamma}$ to our method. We define $\varepsilon_\Gamma$ in Eq.~\ref{eq:wrench_error} as the wrist wrench error of the CLP distribution:
\begin{align}
    \label{eq:wrench_error}
    \varepsilon_{\Gamma} &= \sum_{i=0}^{n_{clp}} s_{i} |\mat{\Gamma}_{i} - \mat{\Gamma}_{EE}|
\end{align}

\noindent $\mathcal{L}_{\Gamma}$ in Eq.~\ref{eq:L_ft} is the summed difference between $\varepsilon_{\Gamma}$ and $\varepsilon_{min}$ for the tool and probe (denoted $_t$ and $_p$ respectively), where $\varepsilon_{min}$ is the lowest error found so far across SCOPE steps for this action:
\begin{align}
    \label{eq:L_ft}
    \mathcal{L}_{\Gamma} &=  \varepsilon_t - \varepsilon_{min,t} + \varepsilon_p - \varepsilon_{min,p}
\end{align}
The inclusion of this loss metric ensures that in addition to the contact, alignment, and penetration constraints, the pose estimates for the objects also explain the experienced wrist wrench. This idea is shown in Fig.~\ref{fig:ft_loss_example}. On the left, the CPFGrasp solution for the given object pose does not resolve the wrist wrench experienced by the robot because the object surface does not contain the ground truth contact location. The object pose on the right is a better estimate because the CPFGrasp solution resolves the wrench that the robot is experiencing at the wrist.  

\subsubsection{Memory Loss}
\label{sec:mem_loss}
\begin{figure}[t]
    \centering
    \includegraphics[width=0.35\textwidth]{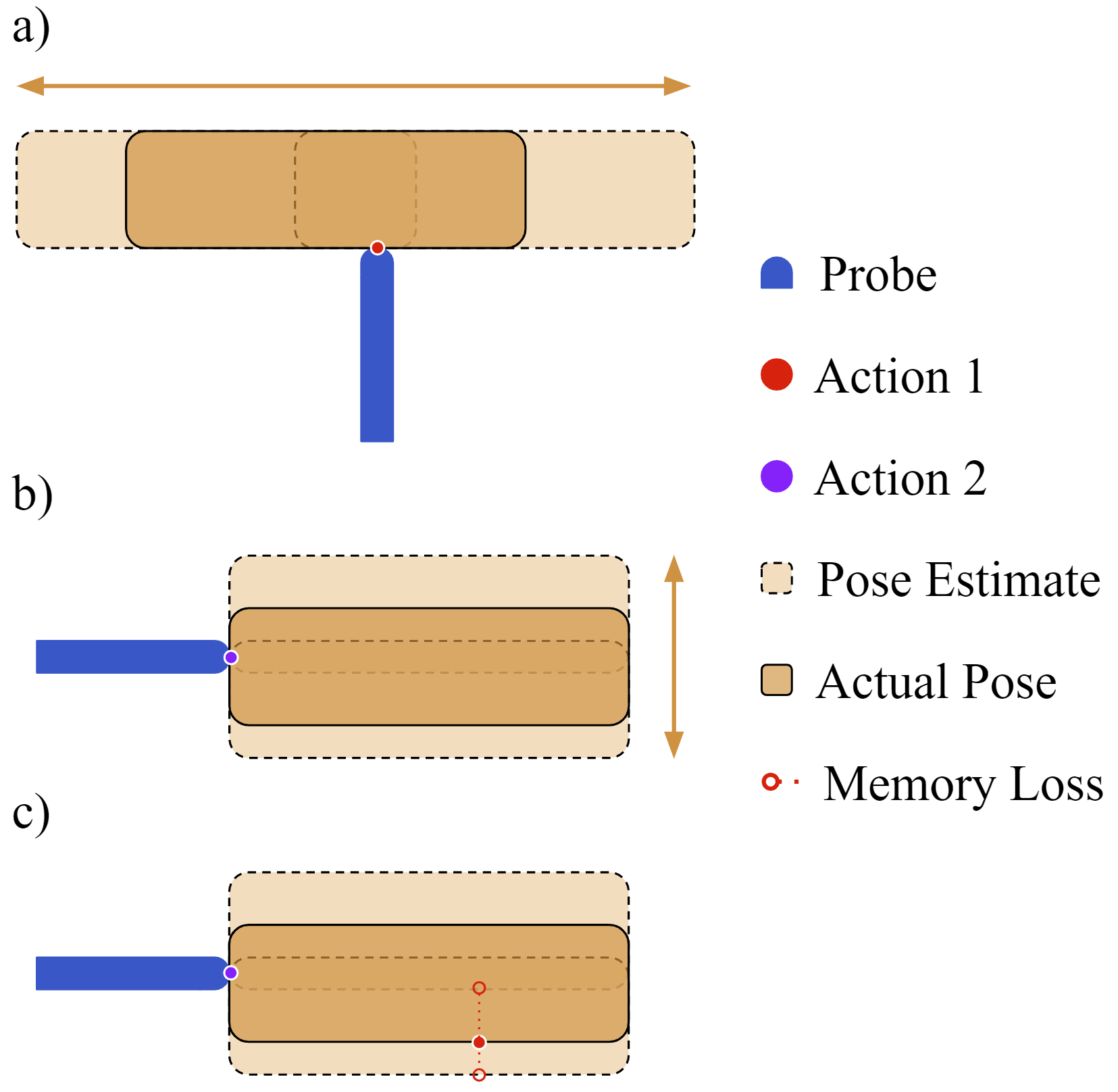}
    \caption{Simple 2-DOF translational example of the intuition behind $\mathcal{L}_M$. a) Probe takes Action 1, resulting in object pose estimates that have horizontal uncertainty. b) Probe takes Action 2, resulting in object pose estimates that have vertical uncertainty. c) By combining Action 1 and Action 2, we can score object pose estimates on how well they satisfy both actions that have been taken and reduce our uncertainty.}
    \label{fig:mem_loss_simple}
\end{figure}

To build MultiSCOPE, we introduce a memory loss $\mathcal{L}_{M}$ to our algorithm. The goal of this loss is to leverage the results of previous interactions and reduce uncertainty in our object pose estimates. This idea is shown in a simple 2-DOF translational case in Fig.~\ref{fig:mem_loss_simple}. The key challenge we address in our implementation is assessing consistency between pose estimates and sequential actions. In order to do this, we introduce the contact clouds $r_{cc}$ for the tool and the probe that contain estimated contact points. The contact clouds grow over time as more actions are taken.

\begin{figure}
    \centering
    \includegraphics[width=0.45\textwidth]{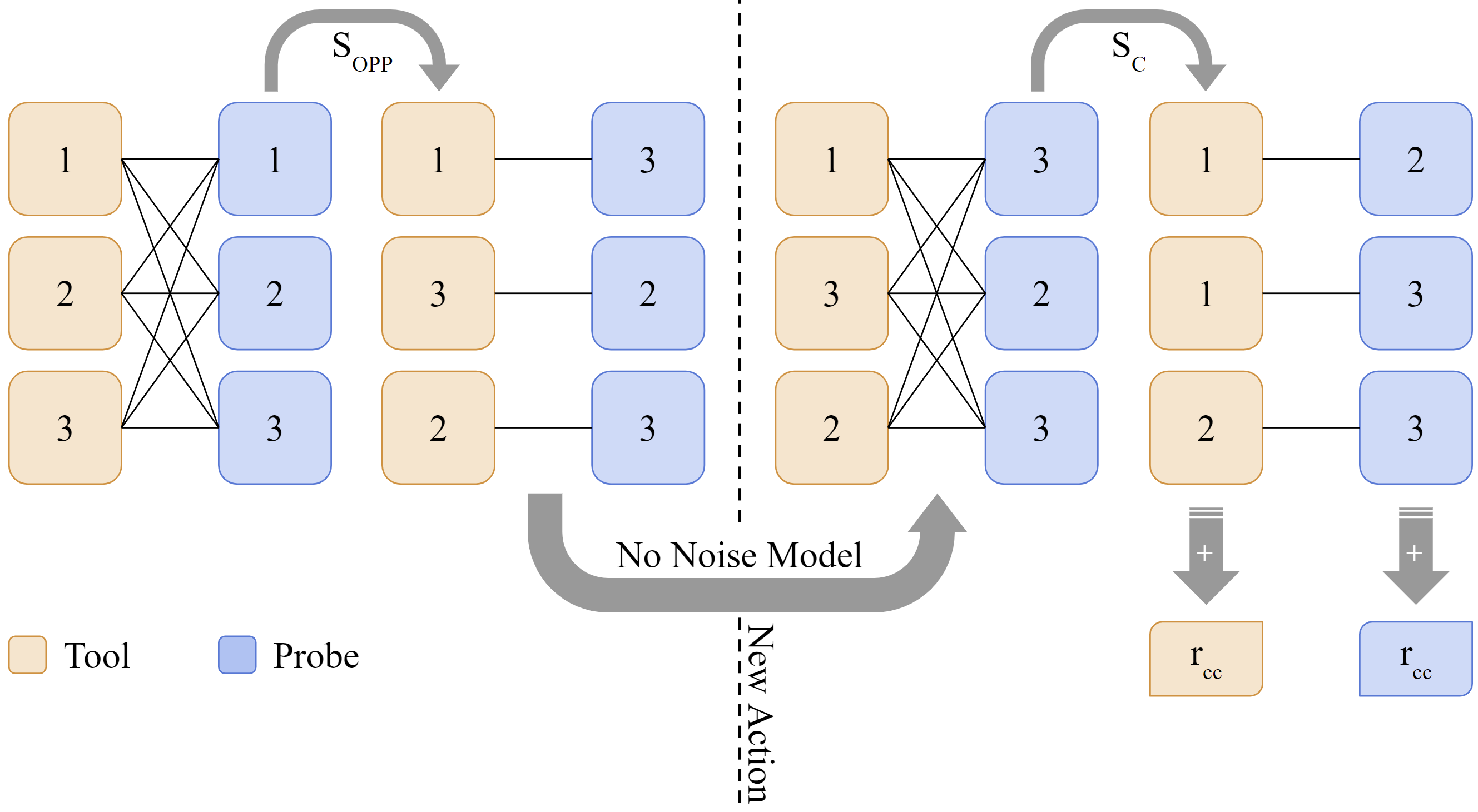}
    \caption{OPPs are represented as blocks, colored orange for the tool and blue for the probe. At the end of one action we use $S_{OPP}$, which includes $\mathcal{L}_M$, to score our top OPP pairs. We then move to the next action without applying the noise model so that we can directly compare the consistency of the last and current OPP pairs using $S_{C}$, which does not include $\mathcal{L}_M$. We take the most consistent particles and use them to update each object's contact cloud $r_{cc}$ for use in memory scoring.}
    \label{fig:cc_update}
\end{figure}

In order to construct the contact cloud with multiple actions, we take the process depicted in Fig.~\ref{fig:cc_update}. We define two scoring functions in Eqs.~\ref{eq:score_consistency} and ~\ref{eq:score_opp}: $S_{OPP}$ and $S_{C}$. 
\begin{align}
    \label{eq:score_consistency}
    S_{C} &= \eta_P \mathcal{L}_{P} + \eta_C \mathcal{L}_{C} + \eta_F \mathcal{L}_{F} + \eta_{\Gamma} \mathcal{L}_{\Gamma} \\
    \label{eq:score_opp}
    S_{OPP} &= \eta_P \mathcal{L}_{P} + \eta_C \mathcal{L}_{C} + \eta_F \mathcal{L}_{F} + \eta_{\Gamma} \mathcal{L}_{\Gamma} + \mathcal{L}_{M}
\end{align}

The difference between these functions is that $S_{C}$ does not include memory loss and is used to directly compare OPP pairs to one another across actions. At the last step of SCOPE for Action $A_{n}$, we find the top $n_{OPP}$ tool and probe pose pairs using $S_{OPP}$. Without applying the noise model to these pairs, we move to Action $A_{n+1}$ and run the first step of SCOPE. Using $S_{C}$ rather than $S_{OPP}$, we again find the top $n_{OPP}$ pairs. By doing this, we are rewarding the OPP pairs that score highly in both $A_{n}$ and $A_{n+1}$ and penalizing the OPP pairs that score well in only $A_{n}$. In order to update the contact cloud, we use Eq.~\ref{eq:rc} to estimate the contact location $\vec{r}_{cc,i}$ for each OPP in the top pairs, then project $\vec{r}_{cc,i}$ back into the frame of $A_{n}$. We weight $\vec{r}_{cc,i}$ by the score of the OPP it belongs to, $S_{c,i}$. For the remaining SCOPE steps in $A_{n+1}$, we use $S_{OPP}$ to take advantage of the contact cloud that we have just updated for each tool. After we update the contact cloud, we project it into the frame of $A_{n+1}$ and use $S_{OPP}$ finish SCOPE for $A_{n+1}$.
\begin{equation}
\label{eq:rc}
    \vec{r}_{cc,i} = S_{C,i} \sum_{j=0}^{N_{clp}} s_{j} \vec{r}_{j}
\end{equation}

When we use the contact cloud $r_{cc}$ to score memory at each step, we compute the weighted average distance from each point in the contact cloud to the surface of the object using a signed distance function (SDF) as shown in Eq.~\ref{eq:mem_loss}.
\begin{equation}
\label{eq:mem_loss}
    \mathcal{L}_M = \sum_{i=0}^{N_{cc,t}} w_{i} SDF_{t}(r_{cc,t_{i}}) + \sum_{i=0}^{N_{cc,p}} w_{i} SDF_{p}(r_{cc,p_{i}})
\end{equation}

We additionally encourage consistency between actions by introducing what we call ``memory dropout''. There are some situations in which none of the OPPs provided by $A_{n}$ are consistent with $A_{n+1}$. We check for this by comparing $S_{C}$ after the last step of $A_{n}$, $S_{C_{-1}}$, with $S_C$ after the first step of $A_{n+1}$, $S_{C_{0}}$. Since the OPPs for each of these steps are the same, we can directly compare their scores. We expect that if the OPPs are consistent between $A_{n}$ and $A_{n+1}$, $S_{C_{0}} \approx S_{C_{-1}}$. If $S_{C_{0}} \gg \Delta_{C} S_{C_{-1}}$, we exclude $A_{n}$ from memory scoring in all future scoring runs and increase the magnitude and spread of the noise model for the rest of $A_{n+1}$.

\subsubsection{Preventing Divergence}
When we inject noise after each SCOPE step, we present the opportunity for the object pose estimate to get worse. Therefore, when it is time to score the OPP pairs we pick the top $n_{OPP}$ object pose pairs between the last and current SCOPE steps.

\begin{algorithm}
\caption{SCOPE: $_{t} \rightarrow$ tool, $_{p} \rightarrow$ probe}
\label{alg:SCOPE}
\begin{algorithmic}
\Procedure{SCOPE}{$\mat{H}, \vec{\Gamma}, f, \vec{\bar{S}}_{C_{-1}}, r_{cc}, n_{A}$}
\State $\mat{H}_{t}, \mat{H}_{p} \gets \mat{H}$
\State $\vec{\Gamma}_{EE,t}, \vec{\Gamma}_{EE,p} \gets \vec{\Gamma}$
\State $f_{t}, f_{p} \gets f$ \Comment{Segmented object faces}
\State $N_{os} \gets $ Number of SCOPE steps
\For{$i \gets 0; \; i < N_{os}; \; i++$}
    \For{$j \gets 0; \; j < N_{opp}; \; j++$}
        \State $(\mat{H}_{t,j}, \mat{H}_{p,j}) \gets \mat{H}_{j}$
        \State $\mat{R}_{c,t} \gets $ CPFGrasp$(f_t, \mat{H}_{t,j}, \vec{\Gamma}_{EE,t})$
        \State $\mat{R}_{c,p} \gets $ CPFGrasp$(f_p, \mat{H}_{p,j}, \vec{\Gamma}_{EE,p})$
    \EndFor
    \If {$i$ is $0$}
        \State $\vec{S}_{OPP} \gets \text{scoreConsistency}(\mat{H}_{t}, \mat{H}_{p}, \mat{R}_{c,t}, \mat{R}_{c,p})$
        \State $r_{cc} \gets$ updateContactCloud($\mat{H}_{t}, \mat{H}_{p}, \mat{R}_{c,t}, \mat{R}_{c,p}$)
        \If {given $\vec{\bar{S}}_{C_{-1}} \textbf{and } \vec{\bar{S}}_{OPP} > \Delta_{C}\vec{\bar{S}}_{C_{-1}}$}
            \State dropMemory($n_{A} - 1$)
        \EndIf
    \Else
        \State $\vec{S}_{OPP} \gets \text{scoreOPPs}(\mat{H}_{t}, \mat{H}_{p}, \mat{R}_{c,t}, \mat{R}_{c,p}, r_{cc})$
        \State $\mat{H} \gets \text{preventDivergence}(\mat{H}$)
    \EndIf
    \If {$i+1 < N_{os}$}
        \State $\mat{H} \gets $ Importance-Resample($\mat{H}, \vec{S}_{OPP}$)
        \State $\mat{H} \gets $ Noise-Model$(\mat{H}, i, n_A)$
    \EndIf
\EndFor
\State \textbf{return} $\mat{H}, \mat{S}$
\EndProcedure
\end{algorithmic}
\end{algorithm}

\section{Experiments}
\label{sec:experiments}

\subsection{Simulation Environment}
\label{sec:sim_env}
We implement our method first in simulation. We used Pybullet \cite{coumans2021} and configured the environment to reflect our hardware setup. We use two Franka Emika Panda robots mounted on the same table. In simulation, both robots use the Panda Hand. We use four tools (wrench, hex key, pawl, and gear) and a probe. We set the ground truth pose to be $[0, 0, 0]_{EE}$ for each tool and the probe. To generate each action, we sample a surface point of the tool and its surface normal. We disable collision between the probe and tool then use the surface point and normal to create a target pose for the probe. To obtain force-torque data, we use Pybullet's applyExternalForce API to apply 3N of normal force to the contact point. We do this because the force-torque generated through contact resolution in Pybullet can be unpredictable and at a different scale from real-world data. We selected 3N of normal force based on similar real-world interactions, where the normal force generally falls between 2N and 5N. We take the joint states of each robot directly from Pybullet as our proprioception data.

\begin{figure}[t]
    \centering
    \includegraphics[width=0.3\textwidth]{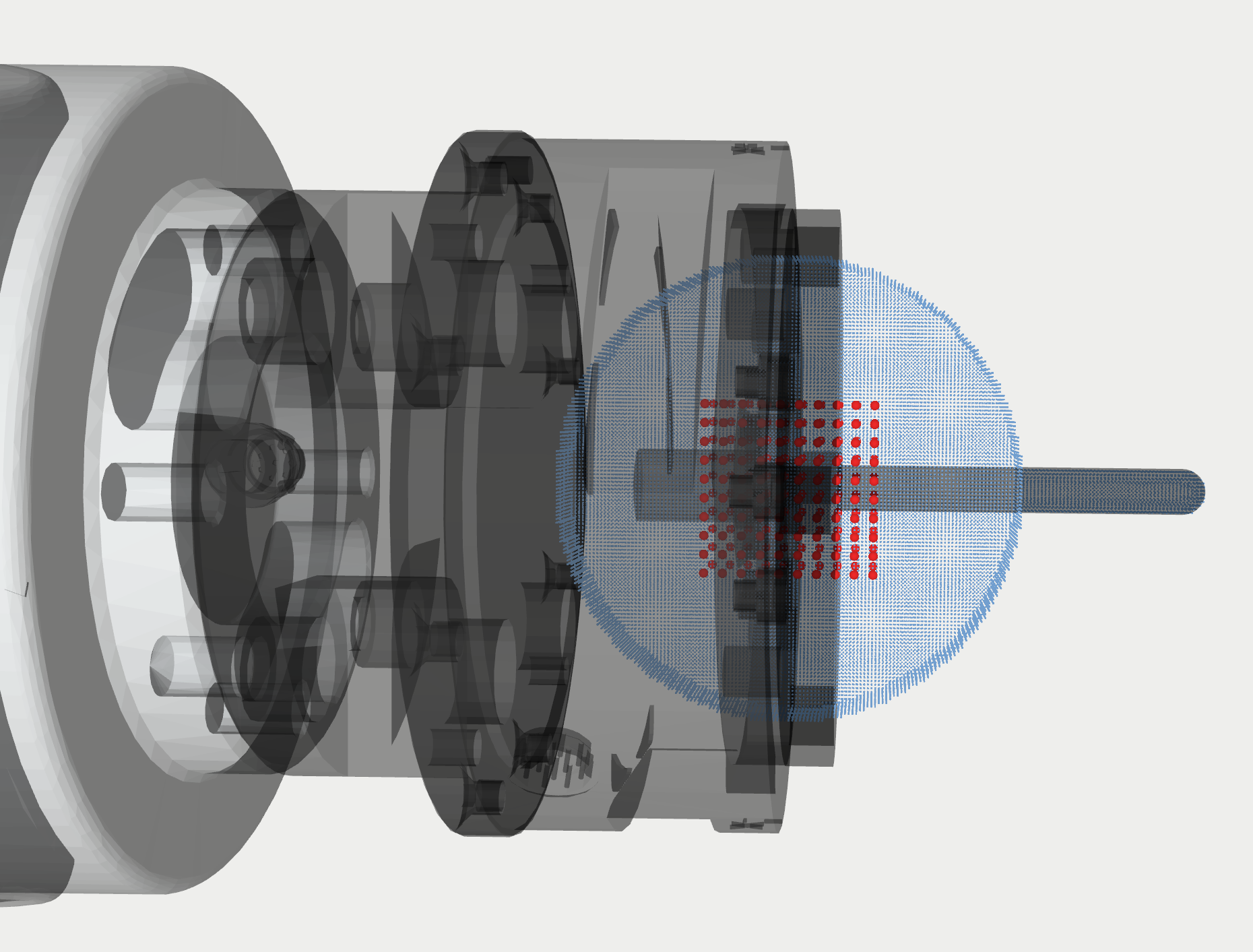}
    \caption{Despite the probe being fixed on the real robot, we treat its pose as being unknown in MultiSCOPE. To do this, we create a graspable probe and use its mesh in MultiSCOPE. The surface points of this mesh are shown in blue. To check grasp validity we create gripper points, shown in red, at the origin of the ATI Gamma FT.}
    \label{fig:rw_probe}
\end{figure}

\subsection{Real-World Environment}
\label{sec:rw_env}
We further test our method in the real world using the wrench tool. We again use two Franka Emika Panda robots mounted on the same table. However, for the real robots we use one robot with a Panda Hand and the other with an ATI Gamma FT sensor mounted at the wrist with the probe attached directly, as shown in Fig.~\ref{fig:teaser}. This shows flexibility in the source of force-torque data. While running MultiSCOPE, we still treat the probe pose as unknown by using the geometry of a graspable probe and simulated gripper fingers at the origin of the Gamma. This is shown in Fig.~\ref{fig:rw_probe}. The 3D-printed wrench tool has a crosshair pattern that interfaces with the Panda Hand fingers to enforce that the ground truth pose is again $[0, 0, 0]_{EE}$. 

One of the assumptions that we make in this method is that we have access to robot proprioception. We do not reason about noise or error in it. However, in the real world, we find that this assumption does not hold to such a high standard. In using our bimanual robot platform, we have found that it is difficult to precisely calibrate the robots with respect to one another even though they are rigidly mounted to the same dimensioned surface. We have developed a simple calibration routine in which we rigidly attach the end-effectors to one another then manually guide the attached arms through the workspace while collecting joint state data. Using this joint state data and geometric information, we map one end-effector frame to the other and create a point cloud for this frame on each robot model. If the robots were perfectly calibrated the point clouds would be identical; however, we have observed that with no calibration we often find 1-1.5 cm of RMS error between the two point clouds. We perform ICP to find the transformation that best maps one point cloud to the other, then apply this transformation to the base of that robot. We find that this method reduces our RMS error to 3-7 mm. 

\begin{figure}[t]
    \centering
    \includegraphics[width=0.4\textwidth]{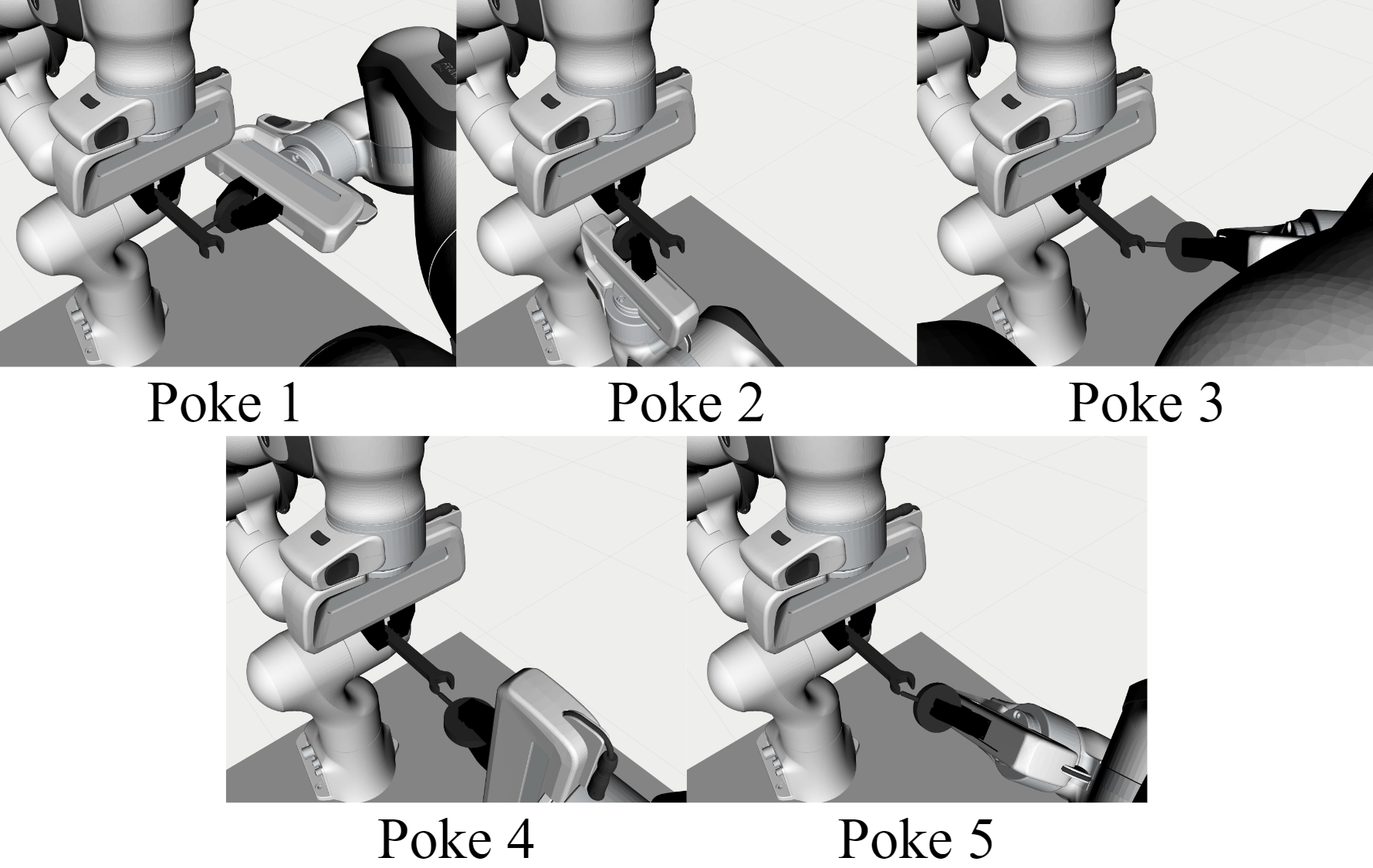}
    \caption{5 poking actions selected for MultiSCOPE trials in simulation.}
    \label{fig:sim_actions}
\end{figure}

In addition to robot proprioception noise and error, we additionally confront force-torque signal noise and error. To address this, we collect static signal before moving into contact in order to zero the sensor and fit the signal noise to a Gaussian distribution. We use the detected noise to form $\Sigma_m$, the calibrated sensor noise used in CPFGrasp. 

\begin{figure}[b]
    \centering
    \includegraphics[width=0.45\textwidth]{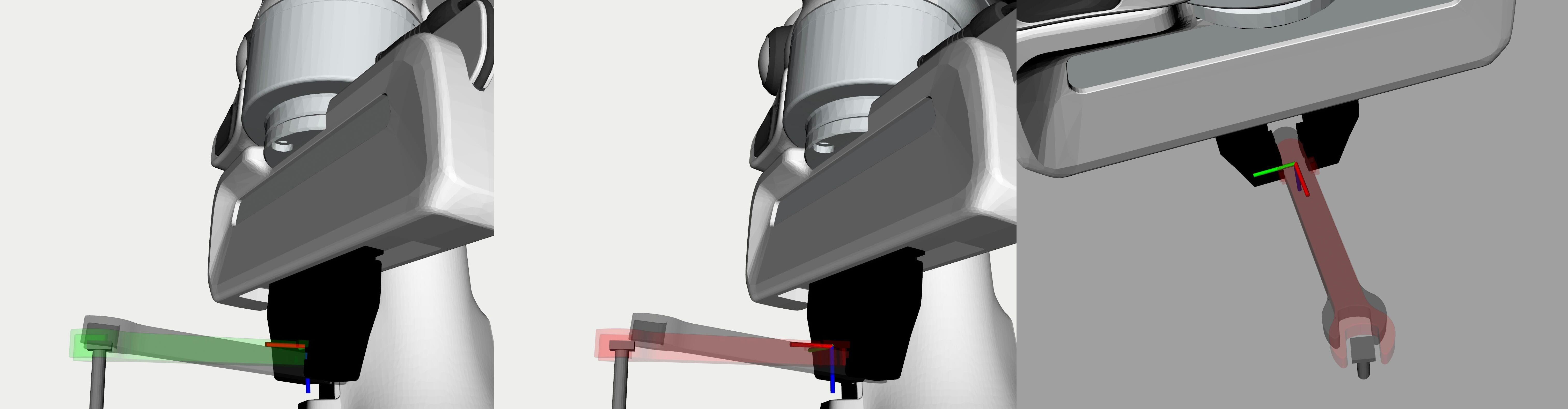}
    \caption{Examples of success and failure cases from the task-based demonstration of the Action 1 results shown in Table~\ref{tab:ablation_action}. The grey wrench tool is the underlying ground truth used for collision checking. In green on the far left is a wrench tool pose estimate that succeeded in surrounding the screw head. The middle and right examples, in red, show wrench tool pose estimates that failed to complete the task. We observe that X-error is very influential in determining task success.}
    \label{fig:action1_wrenchdemo}
\end{figure}

\begin{figure*}
    \centering
    \includegraphics[width=\textwidth]{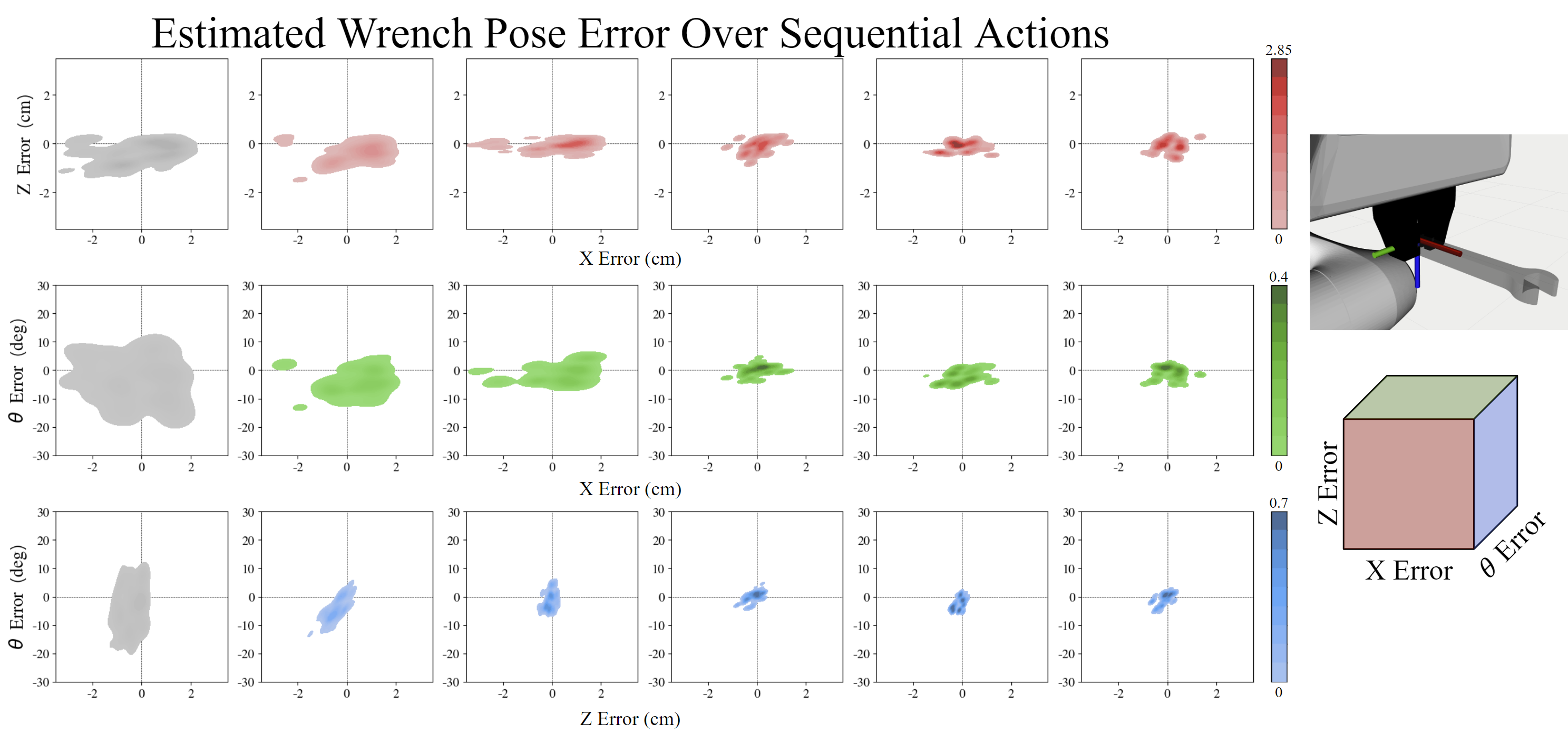}
    \caption{These results are the combined output of all 10 MultiSCOPE trials with the wrench tool. The error is presented in $[x, z, \theta]_{EE}$, which is shown on the upper right with $x$ in red, $y$ in green, and $z$ in blue. We visualize this 3D error as 3 separate planes: X-Z (red), X-$\theta$ (green), and Z-$\theta$ (blue). For clarity, this is shown in the cube on the lower right. The first column (grey) shows the error at initialization and every subsequent column shows the results after taking an additional action. We can see that as MultiSCOPE takes more actions, the pose estimate error shrinks and has higher density about $[0, 0, 0]_{EE}$.}
    \label{fig:sim_results}
\end{figure*}

\subsection{Noise Injection in Simulation}
\label{sec:noise_injection}
To mimic what we observe in real-world hardware, we inject noise into the MultiSCOPE input $\Gamma_{EE} = [F; T] \in \mathbb{R}^{6 \times 1}$ for each arm. For each action, we separately calculate the magnitude of the force $||F||$ and torque $||T||$. We define two Gaussian distributions $\mathcal{N} \sim (0, \sigma)$ where $\sigma_{F}$ is defined as $n$\% of $||F||$ and $\sigma_{T}$ as $n$\% of $||T||$. We modify the MultiSCOPE input to be $\Gamma_{EE,noise} = \Gamma_{EE} + [\delta_{F}; \delta_{T}]$ where $\delta_{F} \sim \mathcal{N}(0, \sigma_{F}) \in \mathbb{R}^{3 \times 1}$ and $\delta_{T} \sim \mathcal{N}(0, \sigma_{T}) \in \mathbb{R}^{3 \times 1}$. Results with the wrench tool for $n = 5\%$ and $n=8\%$ are shown in Sec.~\ref{sec:noise_injection_results}.

\begin{algorithm}
\caption{MultiSCOPE: $_{t} \rightarrow$ tool, $_{p} \rightarrow$ probe}
\label{alg:MultiSCOPE}
\begin{algorithmic}
\Procedure{MultiSCOPE}{$\vec{\Gamma}_{EE,t}, \vec{\Gamma}_{EE,p}$}
\State $\mat{H}_{t} \gets \{\mat{H}_{t, 0}, \cdots, \mat{H}_{t, N_{opp}} \}$
\Comment{Init OPPs}
\State $\mat{H}_{p} \gets \{\mat{H}_{p, 0}, \cdots, \mat{H}_{p, N_{opp}} \}$
\Comment{Init OPPs}
\State $\mat{H} \gets (\mat{H}_{t}, \mat{H}_{p})$
\State $\vec{\Gamma} \gets (\vec{\Gamma}_{EE,t}, \vec{\Gamma}_{EE,p})$
\State $\vec{S} \gets 1/N_{opp}$ \Comment{Initially uniform OPP scores}
\State $r_{cc} \gets \{\}$ \Comment{Initially empty contact cloud} 
\State $N_{A} \gets $ Number of actions
\State $f_{t} \gets \text{segmentTool}(\mathbf{n}_{\mathcal{S},t}, n_{clusters,t}, \varepsilon_t, n_{min,t})$
\State $f_{p} \gets \text{segmentTool}(\mathbf{n}_{\mathcal{S},p}, n_{clusters,p}, \varepsilon_p, n_{min,p})$
\State $f \gets (f_t, f_p)$
\For{$i \gets 0; \; i < N_{A}; \; i++$}
    \If {$i \text{ is } 0$}
        \State $\mat{H}, \mat{S} \gets$ SCOPE($\mat{H}, \vec{\Gamma}, f, r_{cc}, i$)
    \Else
        \State $\mat{H}, \mat{S} \gets$ SCOPE($\mat{H}, \vec{\Gamma}, f, \vec{\bar{S}}, r_{cc}, i$)
    \EndIf
\EndFor
\State \textbf{return} $\mat{H}, \mat{S}$
\EndProcedure
\end{algorithmic}
\end{algorithm}

\subsection{Task-Based Demonstration}
The goal of MultiSCOPE is to estimate in-hand tool poses that are sufficiently accurate to use in manipulation tasks. In order to demonstrate the efficacy of our method, we use the wrench tool to approach a screw in our environment, in preparation to tighten the screw. We assign a planning frame based on the estimated wrench tool pose and plan to an approach point. We then use the Pilz Industrial Motion Planner to plan a linear Cartesian path into contact between the wrench tool and the screw head. We qualify the trial as a success if the wrench tool is able to surround the screw head. Since we are using the ground truth geometry as the collision body, our estimate of wrench tool pose must agree with the ground truth in order to plan successfully. 

We conduct this demonstration in the simulation environment as well as the real world, but with slightly different tolerances. The opening of our wrench tool (which is the 3D printed version of a wrench tool sold by McMaster-Carr) measures $\frac{1}{2}$" (12.7 mm). Therefore, a square-headed bolt must have a minimum side length of 0.35" (8.9 mm) in order to catch in the mouth of the wrench when the wrench is turned. This leaves at most 0.075" (1.9 mm) tolerance on either side of the screw head. We created a mesh for this square-headed screw and added it to our environment for the simulation demonstration. For the real-world demonstration, we increased the maximum tolerance to 3.5 mm on each side by foregoing a screw head and using just a cylinder.

{\renewcommand{\arraystretch}{1.3}
\begin{table*}
\caption{Results of ablation study across loss metrics $\mathcal{L}_P, \mathcal{L}_C, \mathcal{L}_F, \mathcal{L}_{\Gamma},$ and $\mathcal{L}_M$ over 5 trials.}
\label{tab:ablation_loss}
\centering
\begin{tabular}{ c  c  c  c  c  c  c  c  c  c }
    \hline
    \multicolumn{10}{ c }{\textbf{Loss Ablation Study Results}} \\
    \hline
    \multirow{2}{*}{\textbf{Loss}} & \multicolumn{2}{c}{$\mat{H}_{t}$ \textbf{Trans Error (cm)} $\downarrow$}  & \multicolumn{2}{c}{$\mat{H}_{t}$ \textbf{Rot Error (deg)} $\downarrow$}  & \multicolumn{2}{c}{$\mat{H}_{p}$ \textbf{Trans Error (cm)} $\downarrow$}  & \multicolumn{2}{c}{$\mat{H}_{p}$ \textbf{Rot Error (deg)} $\downarrow$} & \multirow{2}{*}{\textbf{Task Success} ($\%$) $\uparrow$} \\
    \cline{2-9}
    & Mean & Stdev & Mean & Stdev & Mean & Stdev & Mean & Stdev & \\
    \hline
    $\mathcal{L}_P$  & 1.32 & 0.74 & 6.55 & 18.26 & 0.89 & 0.35 & 3.67 & 26.77 & 0 \\
    
    $\mathcal{L}_C$ & 0.80 & 0.31 & -0.21 & 2.60 & 0.74 & 0.40 & 2.63 & 5.30 & 40 \\
    
    $\mathcal{L}_F$ & 1.19 & 0.59 & -2.95 & 3.26 & 1.29 & 0.70 & -7.29 & 14.68 & 40 \\
    
    $\mathcal{L}_{\Gamma}$ & 0.46 & 0.27 & -1.26 & 2.10 & 1.06 & 0.64 & -5.36 & 14.83 & 80 \\
    
    $\mathcal{L}_M$ & 1.30 & 0.41 & -1.98 & 2.47 & 1.65 & 0.66 & -6.06 & 20.64 & 20 \\

    $\mathcal{L}_P, \mathcal{L}_C, \mathcal{L}_F$ & 0.66 & 0.37 & -2.13 & 2.64 & 0.82 & 0.44 & 0.08 & 11.63 & N/a \\
    
    $\mathcal{L}_P, \mathcal{L}_C, \mathcal{L}_F, \mathcal{L}_{\Gamma}$ & 0.61 & 0.31 & -0.47 & 2.92 & 0.84 & 0.47 & 2.45 & 11.64 & N/a \\

    $\mathcal{L}_P, \mathcal{L}_C, \mathcal{L}_F, \mathcal{L}_M$ & 0.51 & 0.31 & -1.15 & 2.79 & 0.78 & 0.49 & -0.07 & 11.80 & N/a \\
    \hline
    Full Method & 0.48 & 0.16 & -0.50 & 1.63 & 0.50 & 0.25 & -1.71 & 4.23 & 100 \\
    \hline
\end{tabular}
\end{table*}
}

\section{Results}
\subsection{Simulated Experiments}
For brevity, we present only the results for the wrench tool here. The pose estimation results for the hex key, gear, and pawl tools can be found in Appendix~\ref{sec:appendix}. For the simulated experiments, we generated a set of actions according to Section~\ref{sec:sim_env} and chose 5 of them to be advantageous for reachability. We ran 10 MultiSCOPE trials on this set of actions. We show these 5 actions in Fig~\ref{fig:sim_actions} and present the results of pose estimation and task completion here.

\subsubsection{Pose Estimation}
The robot grasping the wrench tool achieved 0.47 $\pm$ 0.27 cm translation error and -0.76 $\pm$ 2.01${\circ}$ rotation error. For the probe, the robot achieved 0.49 $\pm$ 0.28 cm translation error and -2.67 $\pm$ 4.76$^{\circ}$ rotation error. We visualize the progression of wrench tool pose estimation results over several actions in Fig~\ref{fig:sim_results}.

\begin{figure*}
    \centering
    \includegraphics[width=\textwidth]{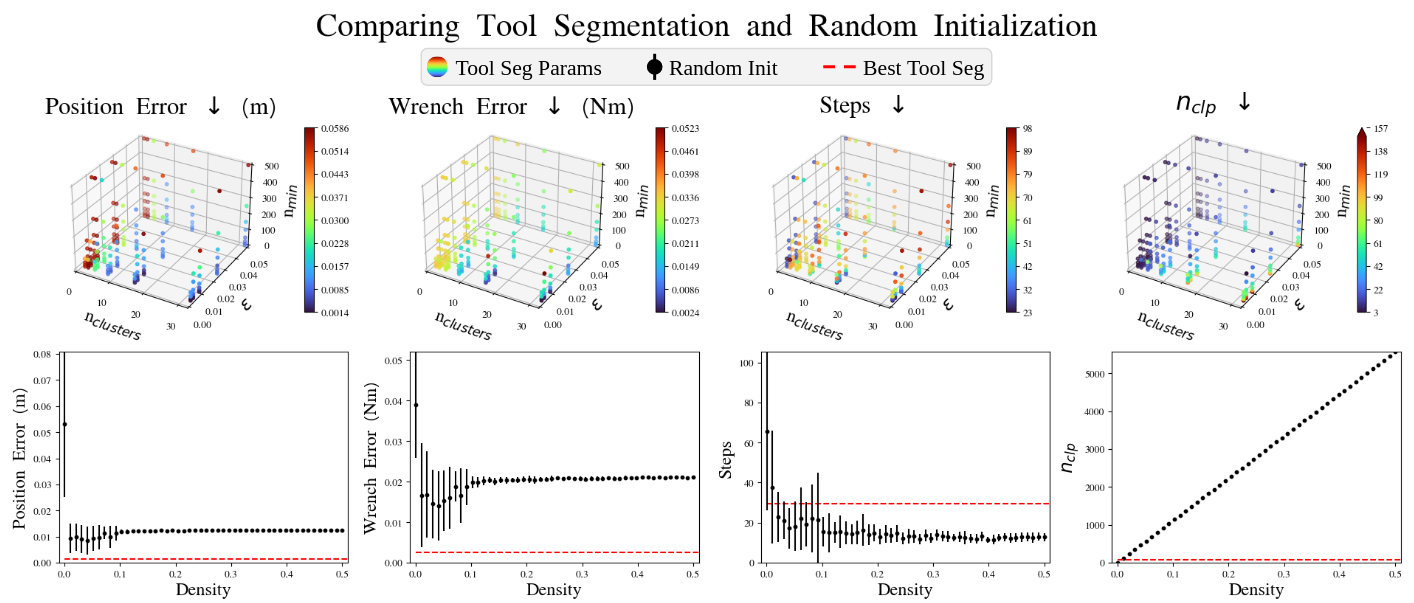}
    \caption{In this figure we compare the results of our tool segmentation algorithm vs. random CLP initialization when using CPFGrasp to estimate contact location with Poke 4 from Fig.~\ref{fig:sim_actions}. We use 4 metrics: the position error, wrench error, number of steps to converge, and the number of CLPs. These results are averaged across 20 random seeds: error bars are shown in the bottom row for random initialization. The 3 parameters for tool segmentation are $n_{clusters}$, $\varepsilon$, and $n_{min}$ while the single parameter for random initialization is the density of the random sampling (fraction of surface points). We find that our tool segmentation method is able to reduce the number of CLPs needed to converge to the ground truth contact location precisely, as shown with red dashed lines.}
    \label{fig:toolseg_comp}
\end{figure*}

\subsubsection{Task Completion} In the task-based demonstration with tight tolerance, the robot succeeded in 8 of 10 trials to surround the screw head with the wrench. In the 2 trials that failed, we found that after Action 1 the pose estimate of the wrench tool maintained relatively large X-error. These poses were maintained in the contact cloud for the tool because they did not meet the qualification for memory dropout mentioned in Sec.~\ref{sec:mem_loss}. Because there was a similar trial that succeeded, we hypothesize that a better-chosen value of $\Delta_{C}$ would improve our success rate.

{\renewcommand{\arraystretch}{1.3}
\begin{table*}
\caption{Results of ablation study across standalone actions over 10 trials.}
\label{tab:ablation_action}
\centering
\begin{tabular}{ c c c c c c c c c c }
    \hline
    \multicolumn{10}{c }{\textbf{Action Ablation Study Results}} \\
    \hline
    \multirow{2}{*}{\textbf{Action}} & \multicolumn{2}{c}{$\mat{H}_{t}$ \textbf{Trans Error (cm)} $\downarrow$}  & \multicolumn{2}{c}{$\mat{H}_{t}$ \textbf{Rot Error (deg)} $\downarrow$}  & \multicolumn{2}{c}{$\mat{H}_{p}$ \textbf{Trans Error (cm)} $\downarrow$}  & \multicolumn{2}{c}{$\mat{H}_{p}$ \textbf{Rot Error (deg)} $\downarrow$}  & \multirow{2}{*}{\textbf{Task Success} ($\%$) $\uparrow$} \\
    \cline{2-9}
    & Mean & Stdev & Mean & Stdev & Mean & Stdev & Mean & Stdev & \\
    \hline
    1 & 1.15 & 0.48 & -3.71 & 3.62 & 0.51 & 0.27 & -2.36 & 5.58 & 20 \\
    2 & 1.72 & 1.11 & -0.23 & 2.46 & 0.58 & 0.23 & -1.79 & 5.59 & 10 \\
    3 & 0.84 & 0.33 & -2.25 & 2.94 & 0.6 & 0.36 & -0.71 & 6.99 & 70 \\
    4 & 0.88 & 0.71 & -2.27 & 2.67 & 0.94 & 0.42 & 2.77 & 9.28 & 50 \\
    5 & 0.83 & 0.46 & -0.86 & 2.56 & 0.75 & 0.30 & 3.51 & 10.93 & 50 \\
    \hline
    Mean & 1.08 & 0.62 & -1.86 & 2.85 & 0.68 & 0.32 & 0.28 & 7.67 & 40 \\
    \hline
    Full Method & 0.47 & 0.27 & -0.76 & 2.01 & 0.49 & 0.28 & -2.67 & 4.76 & 80 \\
    \hline
\end{tabular}
\end{table*}
}

\begin{figure*}
    \centering
    \includegraphics[width=\textwidth]{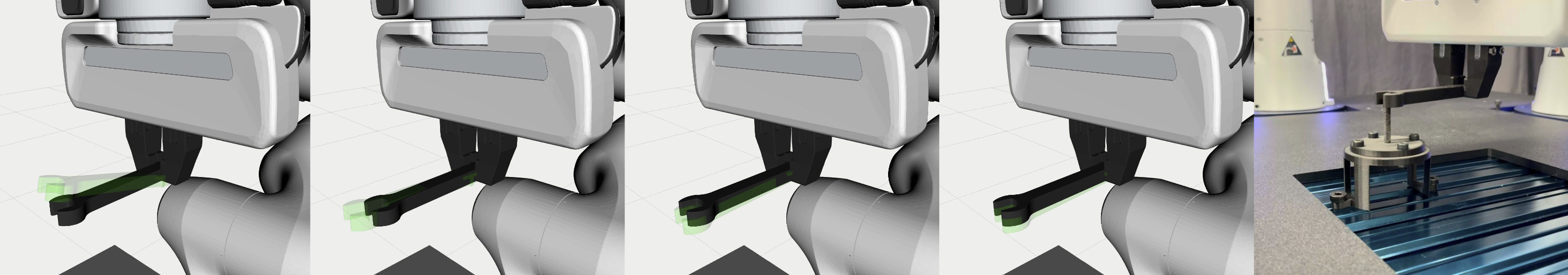}
    \caption{Estimated wrench pose (green, transparent) compared to ground truth (black, opaque) after Action 1 through after Action 4. At the far right, the result of planning using the estimated wrench pose after Action 4 in the real world. The robot succeeded in the task in all 5 real-world trials.}
    \label{fig:hardware_demo}
\end{figure*}

\subsubsection{Noise Injection}
\label{sec:noise_injection_results}
We sampled force-torque noise at 5\% and 8\% of each action's magnitude according to Sec.~\ref{sec:noise_injection} and tested in 5 trials. At 5\% noise, we found that the robot grasping the wrench tool achieved 0.52 $\pm$ 0.31 cm average translation error and -0.1 $\pm$ 3.08$^{\circ}$ rotation error. The robot grasping the probe achieved 0.58 $\pm$ 0.48 cm of translation error and 0.76 $\pm$ 5.38$^{\circ}$ of rotation error. Additionally, the robot was able to complete 4 of 5 task-based demonstrations. At 8\% of each action's magnitude, we found that MultiSCOPE was no longer able to achieve results that were sufficiently accurate to complete any demonstrations successfully. The results at this 8\% noise were 1.13 $\pm$ 0.33 cm translation error and 0.87 $\pm$ 1.27$^{\circ}$ rotation error for the wrench tool and 0.94 $\pm$ 0.62 cm translation error and -1.31 $\pm$ 5.09$^{\circ}$ rotation error for the probe. 

\subsubsection{Ablation Studies}
We conducted several ablation studies to examine how our selected loss metrics impacted our results. In addition to examining the effect of each loss metric by itself, we also compared the results of performing only a single action. We present these results both in terms of their $[x, z, \theta]_{EE}$ errors as well as their task success rate. Results of these ablation studies are presented in Table~\ref{tab:ablation_loss} and Table~\ref{tab:ablation_action}. 

\subsection{Hardware Demonstrations}
For the real robot demonstration, we chose and collected data for 4 actions and ran 5 MultiSCOPE trials. 3 of the 4 actions are shown in Fig.~\ref{fig:teaser}. The last action is in the same direction as Poke 3 on a different face of the wrench. We present the results of the task-based demonstration trials in Fig.~\ref{fig:hardware_demo}. We successfully completed the task in 5 of 5 trials despite the robot proprioception error discussed in Sec.~\ref{sec:rw_env}.

\section{Conclusion}

In this paper, we demonstrated the efficacy of MultiSCOPE in disambiguating in-hand object poses using robot proprioception and tactile feedback. We evaluated our method in both simulated and real world environments using Franka Emika Panda robots and saw improved performance from using multiple actions rather than single actions. These improvements enabled us to use the grasped tool for task-based demonstrations. 

Building on the success of this work, we look ahead to more principled action selection to reduce the number of actions needed to converge to accurate object pose estimates. We also consider integrating vision or visuotactile sensing as relevant extensions to our method to increase efficiency and applicability. 

\bibliographystyle{plainnat}
\bibliography{ref}

\begin{appendices}
\section{Extended Results}
\label{sec:appendix}
\begin{figure}

    \subfloat[Hex Key Actions]{%
      \includegraphics[clip,width=\columnwidth]{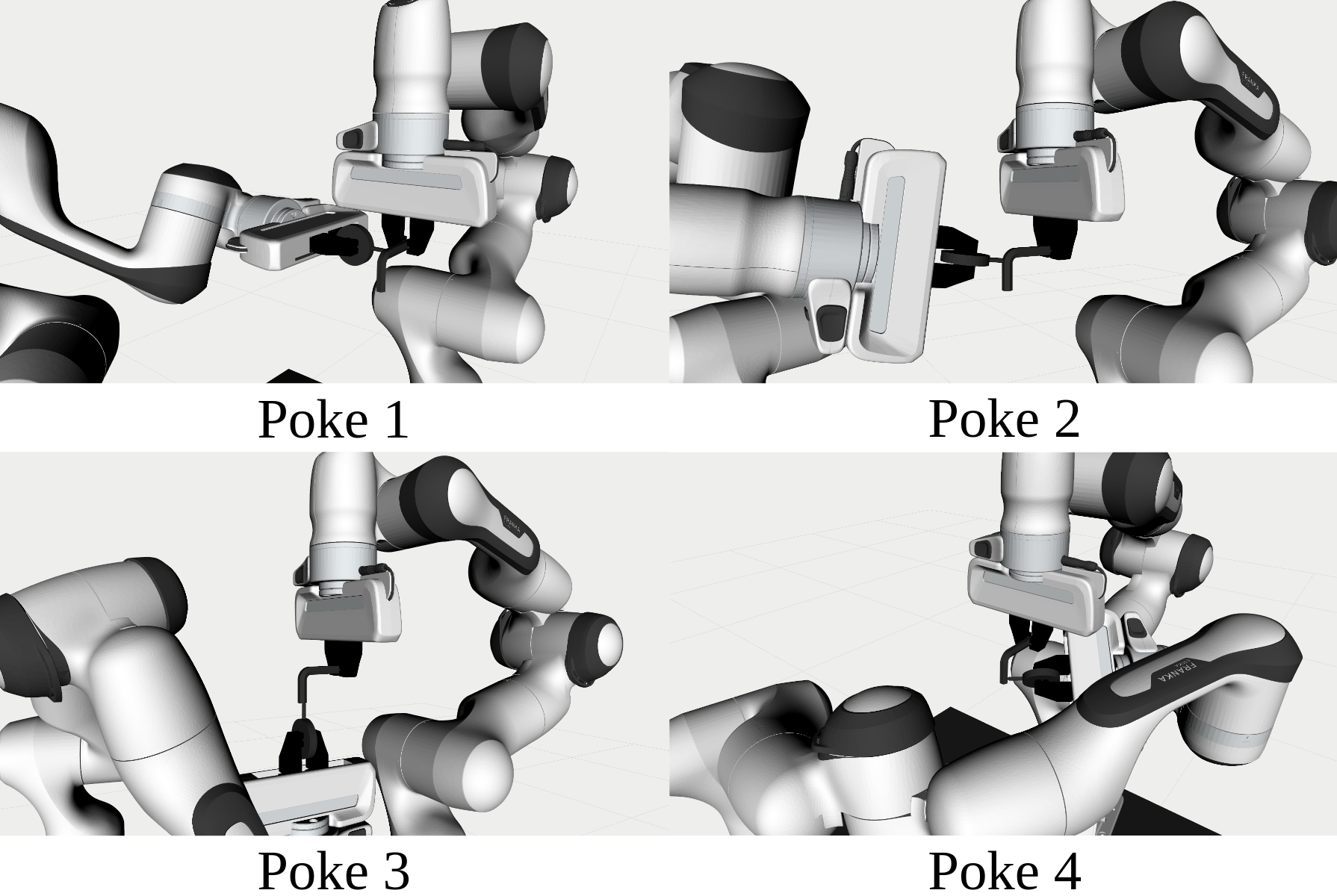}%
    }
    
    \subfloat[Pawl Actions]{%
      \includegraphics[clip,width=\columnwidth]{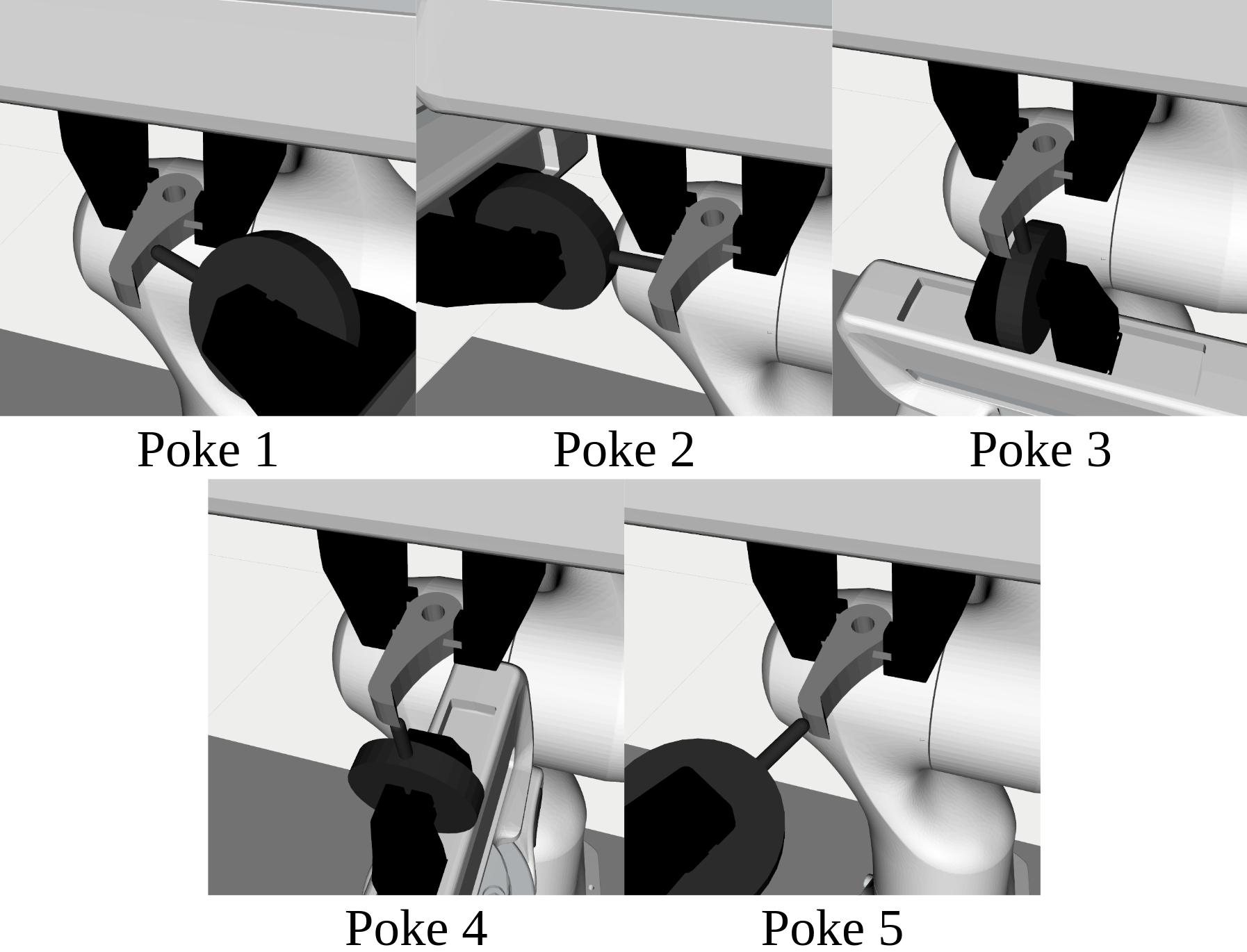}%
    }
    
    \subfloat[Gear Actions]{%
      \includegraphics[clip,width=\columnwidth]{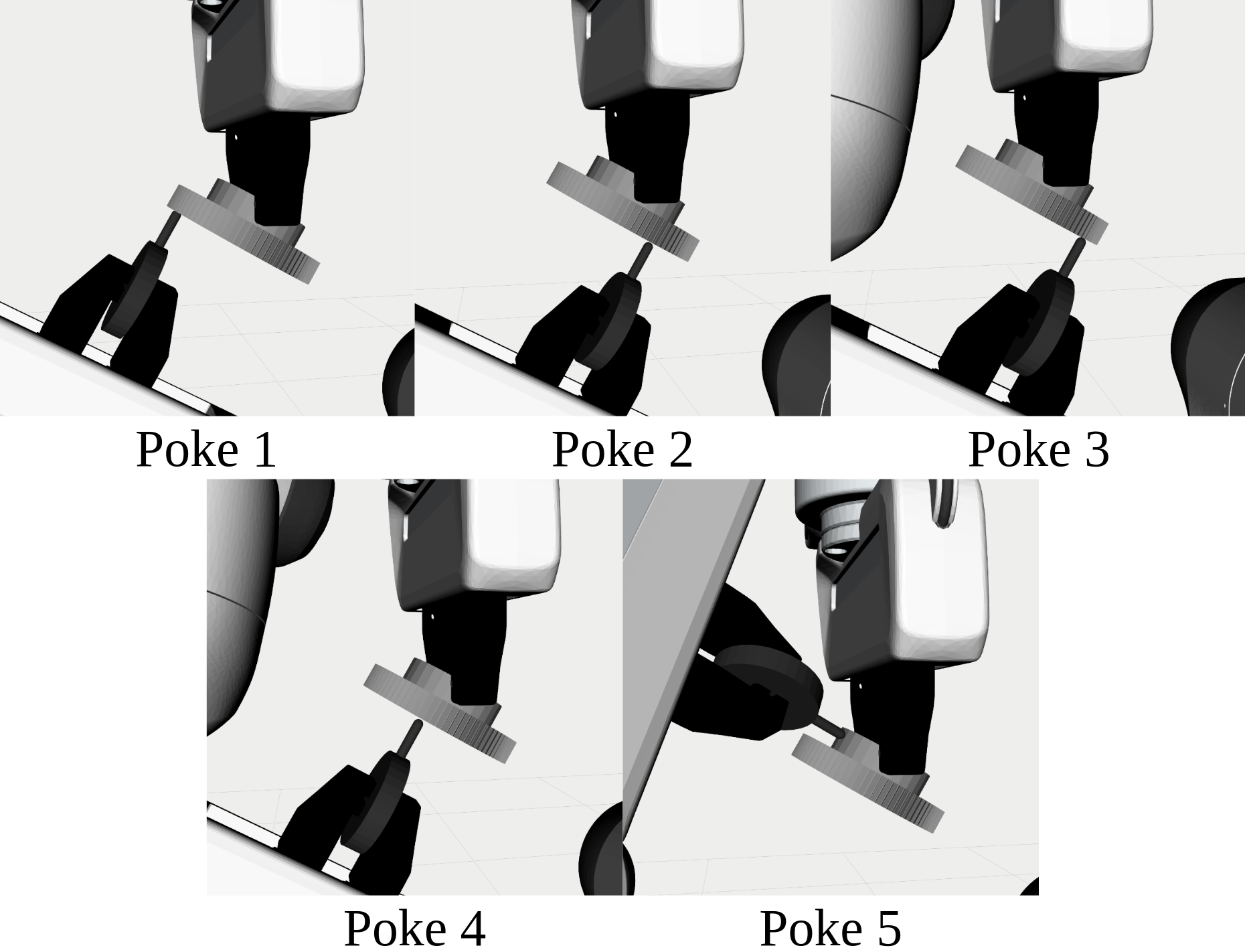}%
    }
    
    \caption{Selected poking actions for MultiSCOPE trials in simulation with the hex key, pawl, and gear tools.}
    \label{fig:actions}
\end{figure}

\begin{figure*}[h]
    \centering
    \includegraphics[width=\textwidth]{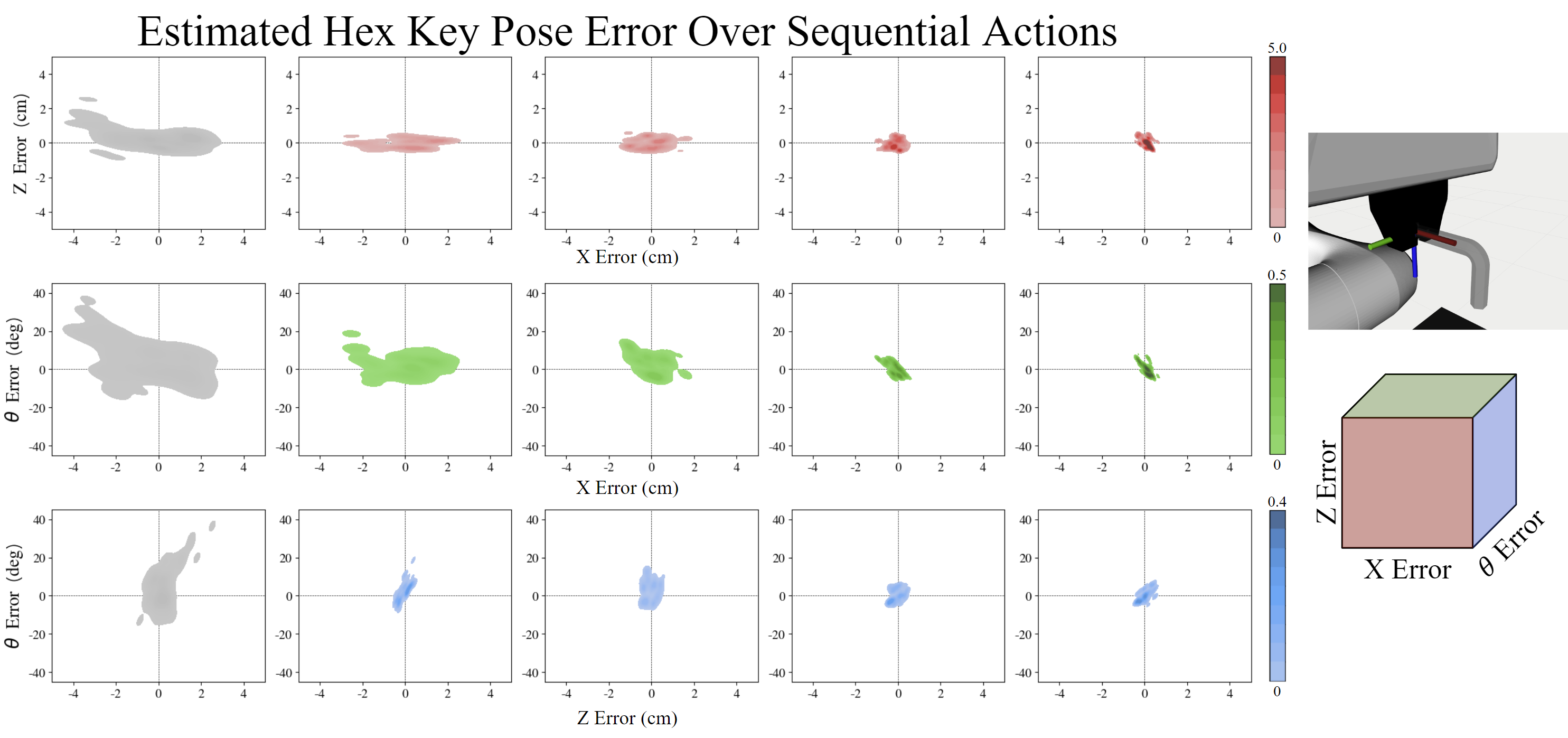}
    \caption{These results are the combined output of all 10 MultiSCOPE trials with the hex key tool. The error is presented in $[x, z, \theta]_{EE}$, which is shown on the upper right with $x$ in red, $y$ in green, and $z$ in blue. We visualize this 3D error as 3 separate planes: X-Z (red), X-$\theta$ (green), and Z-$\theta$ (blue). For clarity, this is shown in the cube on the lower right. The first column (grey) shows the error at initialization and every subsequent column shows the results after taking an additional action.}
    \label{fig:hex_results}
\end{figure*}

\begin{figure*}[h]
    \centering
    \includegraphics[width=\textwidth]{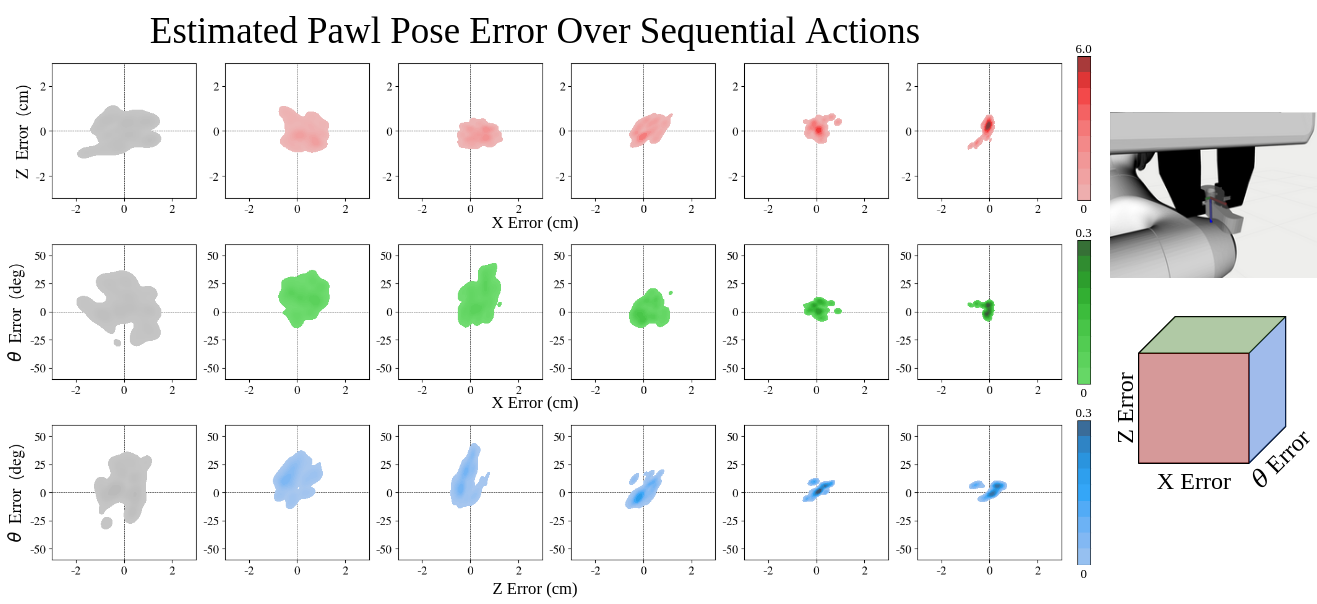}
    \caption{These results are the combined output of all 10 MultiSCOPE trials with the pawl tool. The error is presented in $[x, z, \theta]_{EE}$, which is shown on the upper right with $x$ in red, $y$ in green, and $z$ in blue. We visualize this 3D error as 3 separate planes: X-Z (red), X-$\theta$ (green), and Z-$\theta$ (blue). For clarity, this is shown in the cube on the lower right. The first column (grey) shows the error at initialization and every subsequent column shows the results after taking an additional action.}
    \label{fig:pawl_results}
\end{figure*}

\begin{figure*}[h]
    \centering
    \includegraphics[width=\textwidth]{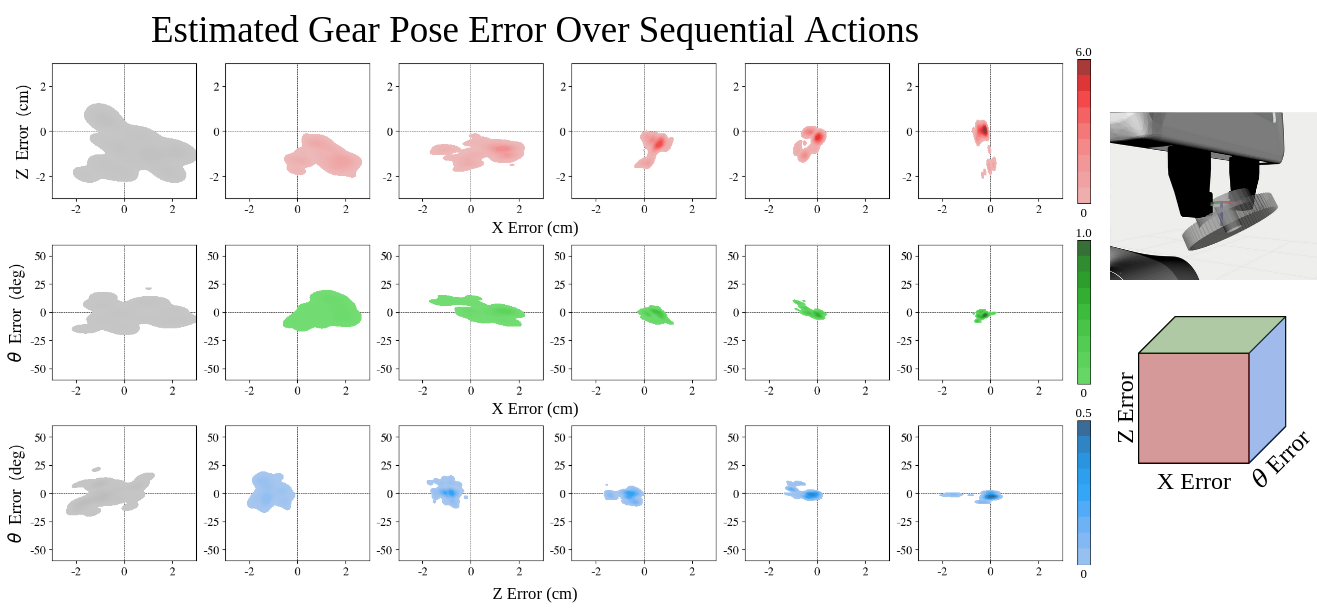}
    \caption{These results are the combined output of all 10 MultiSCOPE trials with the gear tool. The error is presented in $[x, z, \theta]_{EE}$, which is shown on the upper right with $x$ in red, $y$ in green, and $z$ in blue. We visualize this 3D error as 3 separate planes: X-Z (red), X-$\theta$ (green), and Z-$\theta$ (blue). For clarity, this is shown in the cube on the lower right. The first column (grey) shows the error at initialization and every subsequent column shows the results after taking an additional action.}
    \label{fig:gear_results}
\end{figure*}

For brevity, we excluded these results from the main body of the paper. We demonstrate that MultiSCOPE is effective at disambiguating in-hand object poses for three additional tools: a hex key, pawl, and gear. We take the simulated actions shown in Fig.~\ref{fig:actions} and conduct 10 trials with each tool. These results are compared to the MultiSCOPE results with the wrench tool in Table~\ref{tab:tool_results_comparison}. We find that the results are consistent across tools. We visualize the pose estimation results for the additional tools in Figs.~\ref{fig:hex_results}, ~\ref{fig:pawl_results}, and ~\ref{fig:gear_results} in the same style as Fig.~\ref{fig:sim_results}.

{\renewcommand{\arraystretch}{1.3}
\begin{table*}
\caption{Results of MultiSCOPE with different tools over 10 trials.}
\label{tab:tool_results_comparison}
\centering
\begin{tabular}{ c c c c c c c c c }
    \hline
    \multicolumn{9}{c }{\textbf{MultiSCOPE Results}} \\
    \hline
    \multirow{2}{*}{\textbf{Tool}} & \multicolumn{2}{c}{$\mat{H}_{t}$ \textbf{Trans Error (cm)} $\downarrow$}  & \multicolumn{2}{c}{$\mat{H}_{t}$ \textbf{Rot Error (deg)} $\downarrow$}  & \multicolumn{2}{c}{$\mat{H}_{p}$ \textbf{Trans Error (cm)} $\downarrow$}  & \multicolumn{2}{c}{$\mat{H}_{p}$ \textbf{Rot Error (deg)} $\downarrow$} \\
    \cline{2-9}
    & Mean & Stdev & Mean & Stdev & Mean & Stdev & Mean & Stdev \\
    \hline
    Wrench & 0.47 & 0.27 & -0.76 & 2.01 & 0.49 & 0.28 & -2.67 & 4.76 \\
    \hline
    Hex Key & 0.31 & 0.16 & 0.00 & 3.03 & 0.30 & 0.18 & 2.29 & 3.60 \\
    \hline
    Pawl & 0.33 & 0.22 & 2.23 & 4.25 & 0.63 & 0.57 & -2.48 & 10.96 \\
    \hline
    Gear & 0.49 & 0.40 & -2.66 & 2.49 & 1.10 & 0.57 & -3.48 & 12.34 \\
    \hline
\end{tabular}
\end{table*}
}

\end{appendices}

\end{document}